%% file: main.tex
\documentclass{vgtc}                          

\graphicspath{{figures/}{pictures/}{images/}{./}} 

\usepackage{times}                     

\usepackage{tabu}                      
\usepackage{booktabs}                  
\usepackage{lipsum}                    
\usepackage{mwe}                       

\usepackage{mathptmx}                  

\usepackage{stfloats}
\usepackage{enumitem}

\onlineid{0}

\vgtccategory{Research}

\vgtcinsertpkg

\title{Evaluating the Semantic Profiling Abilities of LLMs\\ for Natural Language Utterances in Data Visualization}

\author{Hannah K. Bako\thanks{e-mail: hbako@cs.umd.edu}\\ %
        \scriptsize University of Maryland%
\and Arshnoor Bhutani\thanks{e-mail: arshnoor@terpmail.umd.edu}\\ %
     \scriptsize University of Maryland %
\and Xinyi Liu\thanks{e-mail: xinyi.liu@utexas.edu}\\ %
     \scriptsize University of Texas at Austin %
\and
 Kwesi A. Cobbina\thanks{e-mail: kcobbina@cs.umd.edu}\\ %
     \scriptsize University of Maryland %
\and Zhicheng Liu\thanks{e-mail: zcliu@cs.umd.edu}\\ %
     \parbox{1.4in}{\scriptsize \centering University of Mayland}}

\abstract{
\input{Sections/abstract}
} 

\CCScatlist{
    \CCScatTwelve{Human-centered computing}{Visu\-al\-iza\-tion}{Empirical studies in visualization}{};
}

\input{commands}
\begin{document}

\firstsection{Introduction}
\maketitle

\input{Sections/introduction}
\input{Sections/related-works}

\input{Sections/ground-truths}

\input{Sections/experiment}

\input{Sections/results}
\input{Sections/discussion}
\acknowledgments{
Thanks to the Human-Data Interaction Group for their feedback and support. This work was supported by NSF grant IIS-2239130.}

\bibliographystyle{abbrv-doi-hyperref}

\bibliography{references}
\end{document}

%% file: Sections/abstract.tex
Automatically generating data visualizations in response to human utterances on datasets necessitates a deep semantic understanding of the data utterance, including implicit and explicit references to data attributes, visualization tasks, and necessary data preparation steps. Natural Language Interfaces (NLIs) for data visualization have explored ways to infer such information, yet challenges persist due to inherent uncertainty in human speech. Recent advances in Large Language Models (LLMs) provide an avenue to address these challenges, but their ability to extract the relevant semantic information remains unexplored. In this study, we evaluate four publicly available LLMs (GPT-4, Gemini-Pro, Llama3, and Mixtral), investigating their ability to comprehend utterances even in the presence of uncertainty and identify the relevant data context and visual tasks. Our findings reveal that LLMs are sensitive to uncertainties in utterances. Despite this sensitivity, they are able to extract the relevant data context. However, LLMs struggle with inferring visualization tasks. Based on these results, we highlight future research directions on using LLMs for visualization generation. Our supplementary materials have been shared on GitHub: \url{https://github.com/hdi-umd/Semantic_Profiling_LLM_Evaluation}.

%% file: commands.tex
\newcommand{\mixtral}{%
  \begingroup\normalfont
  \includegraphics[height=\fontcharht\font`\B]{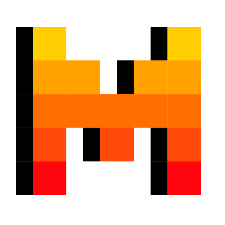}%
  \endgroup
}

\newcommand{\llama}{%
  \begingroup\normalfont
  \includegraphics[height=\fontcharht\font`\B]{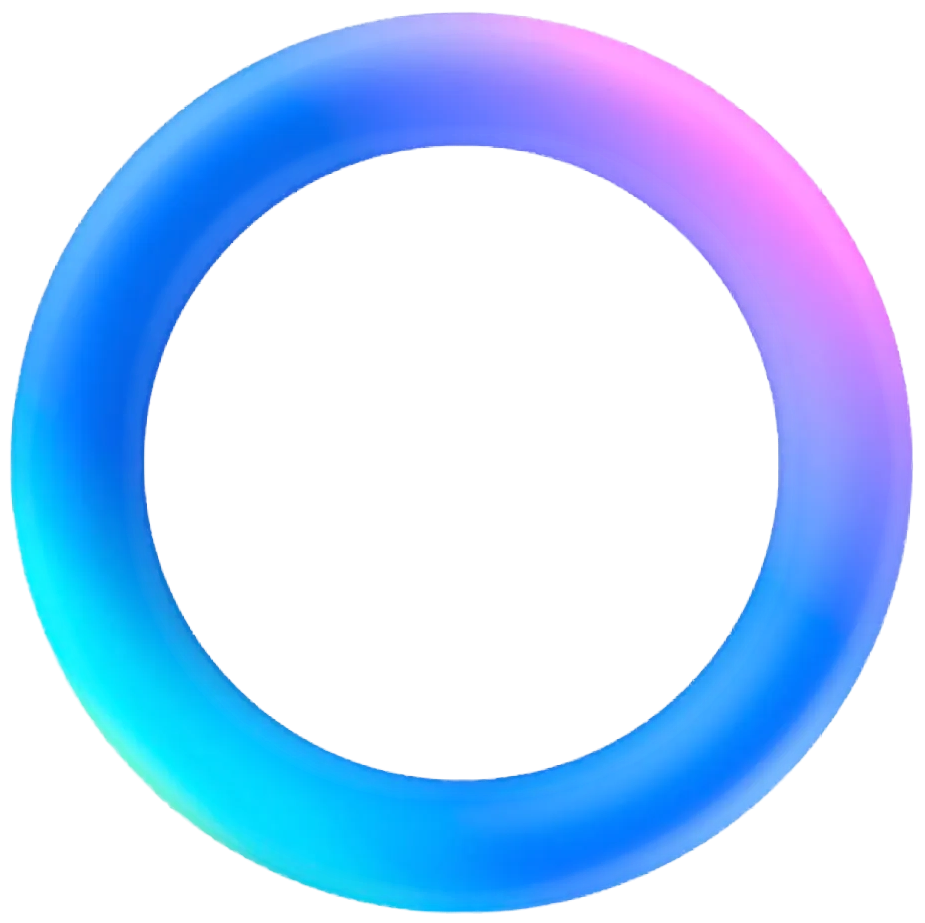}%
  \endgroup
}
\newcommand{\gemini}{%
  \begingroup\normalfont
  \includegraphics[height=\fontcharht\font`\B]{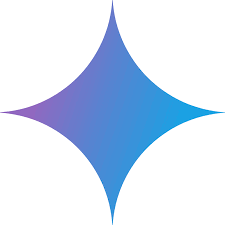}%
  \endgroup
}

\newcommand{\gpt}{%
  \begingroup\normalfont
  \includegraphics[height=\fontcharht\font`\B]{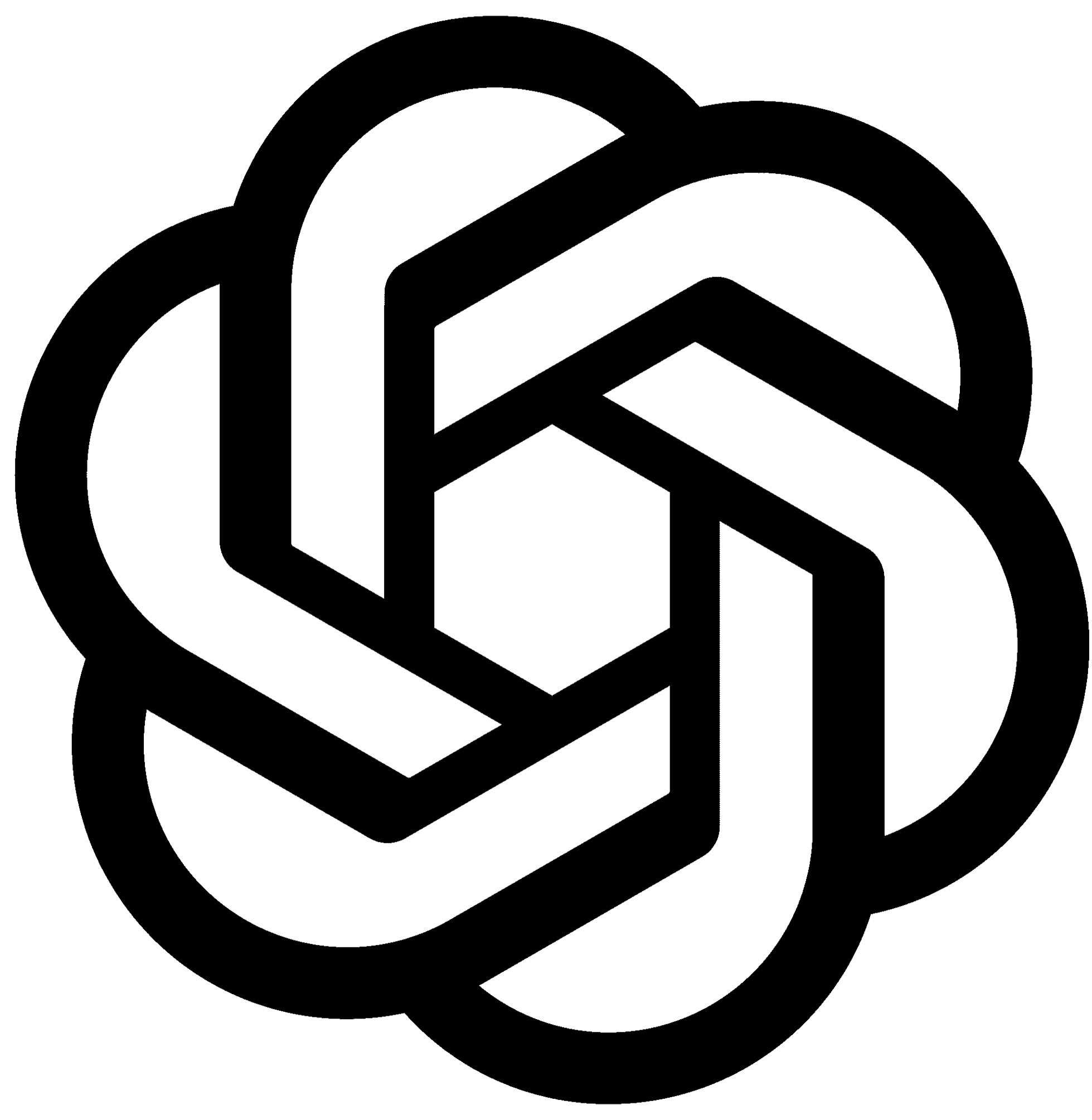}%
  \endgroup
}

\newcommand{\revision}[1]{
\textcolor{black}{#1}
}

\newif\ifnotes
\notestrue
\definecolor{lilac}{RGB}{177,156,217}

\usepackage{multirow}
\usepackage{enumitem}
\usepackage{booktabs}
\usepackage{hyperref} 

\usepackage{caption}
\usepackage{subcaption}

%% file: Sections/introduction.tex
Designing an effective data visualization requires multiple considerations, such as identifying relevant data attributes, preparing the dataset in the right format through data wrangling and transformation, identifying the analytic tasks or communication goals, and choosing appropriate visual encoding strategies. 
Over the years, visualization researchers have primarily focused on different ways to automatically identify appropriate visual encodings 
~\cite{wongsuphasawat2016voyager, mackinlay2007showme, zhao2020chartseer}, but have largely overlooked important aspects such as automating task identification and data preparation.
Only recently have researchers started to address these overlooked issues~\cite{wu2022NL2Vis, narechania2021nl4dv,wang2021falx}.


Among these efforts, natural language interfaces (NLI) have emerged as a popular interaction paradigm for visualization generation. To users, it is easier to articulate their visualization intents through natural language than using programming constructs or complex graphical user interfaces; to system builders, natural language utterances provide valuable information on user intent that could be hard to capture. 
However, natural language utterances can be difficult to handle due to uncertainties such as ambiguities \cite{gao2015datatone} and under-specification~\cite{setlur2019inferencing}. Furthermore, it is necessary to address issues such as data preparation and task identification in visualization systems with natural language interfaces.


Large Language Models (LLMs) are promising in providing a foundation for natural language interfaces tailored to data visualization, 
due to their ability to interpret and generate textual data. 
While a few tools have utilized them for visualization generation~\cite{tian2024chartgpt, han2023chartllama, wang2024dataformulator, dibia2023lida},  
they tend to focus on low-level applications of LLMs, such as generating code for data transformations~\cite{wang2024dataformulator} or simply integrating them as part of a pipeline~\cite{dibia2023lida}.
It is still unclear how well LLMs perform at extracting information crucial to visualization generation from utterances without human interference. 

In this work, we embark on an evaluation of the capabilities of LLMs in the \textbf{semantic profiling} of natural language utterances for \revision{the purpose of} data visualization generation. In line with other work, we use the term ``utterance'' to refer to questions or instructions people use to elicit responses from an NLI or LLM~\cite{srinivasan2021collecting}. 
By semantic profiling, we do not evaluate visualizations generated by LLMs but instead focus on the following dimensions: 1) \textit{clarity analysis}, which determines if an utterance is ambiguous, under-specified, or asking for missing data, 2) \textit{data attribute and transformation identification}, which identifies relevant data columns and any necessary transforms to prepare the data into a usable format, and 3) \textit{task classification}, which seeks to uncover user intent. 



To support our research goal, we collated a corpus of 500 data-related utterances based on an evaluation of two NL datasets (NLVCorpus~\cite{srinivasan2021collecting} and Quda~\cite{fu2020quda}). We analyzed utterances with the following annotations: 1) uncertainties such as ambiguities and missing data references, 2) required data attributes and data transformations, and 3) visualization tasks. We then present a systematic analysis of the capabilities of four publicly available LLMs (GPT-4, Llamma3, Mixtral, and Gemini) across the three dimensions of semantic profiling.  Our results show that LLMs make inferences at a different level of abstraction than humans, causing them to be hyper-sensitive to uncertainties in utterances. We also find that LLMs perform reasonably at identifying the relevant data columns and data transformations expressed in utterances but are not able to properly infer visualization tasks. We highlight our observations on the current strengths and challenges of LLMs and present a discussion on considerations for using LLMs in visualization generation.

%% file: Sections/related-works.tex
\section{Related Work}
\noindent\textbf{Natural Language Interfaces for Visualization Generation.} There has been extensive research on natural language interfaces (NLI) dating as far back as 2001 when Cox et al. proposed the use of natural language as an input medium for the generation of data visualizations~\cite{cox2001multi}. Since then, a plethora of NLIs have been created~\cite{gao2015datatone,sun2010articulate,kassel2018valletto, hoque2018pragmatics, kumar2016towards, narechania2021nl4dv}. These NLIs use techniques, such as lexical tokenization or semantic parsing, to infer and translate representations of data attributes and tasks in utterances into visualizations. However, when users' utterances are under-specified, inferring the correct data and task representation becomes challenging.
Tools such as DataTone circumvent this limitation by 
allowing users to resolve ambiguity through GUI widgets. Similarly, Eviza~\cite{setlur2016eviza} and Evizeon~\cite{hoque2018pragmatics} provide users with the ability to interact with generated visualizations and refine designs via follow-up utterances.

Recent research has progressed towards facilitating visualization code generation based on NL input~\cite{wu2022NL2Vis,narechania2021nl4dv}
, generating NL explanations for visualizations~\cite{kim2020answering} and recommending input utterances~\cite{srinivasan2021snowy}. Together, these works demonstrate the capabilities of NLIs for visualization. \revision{However, NLIs still struggle with resolving under-specifications in utterances without human intervention.}

\vspace{0.5mm}
\noindent\textbf{Large Language Models for Data Visualization.}
Technological advances have given rise to improvements in NLIs, such as the use of BERT to translate user intent expressed in NL into
a domain-specific language for visualizations~\cite{chen2022type}. More recently, we have seen an uptick in the applications of Large Language Models for visualization generation.
\revision{One such tool is ChartLlama~\cite{han2023chartllama}, which uses a fine-tuned open-source LLM trained on synthetic benchmark dataset generated from GPT-4~\cite{openaigpt42023} to}
enhance chart generation and comprehension. Some tools develop pipelines to prompt LLM for relevant code for visualization implementations~\cite{tian2024chartgpt, dibia2023lida, maddigan2023chat2vis}, while others use  LLMs to facilitate data transformations~\cite{wang2024dataformulator}.

There have also been works that evaluate the capabilities of LLMs for different visualization \revision{contexts}. Li et al. evaluate prompting strategies for generating visualizations based on the nvbench dataset~\cite{li2024visualization}. 
Vázquez also evaluates LLMs across 3 axes: the variety of generated chart types, supported libraries, and design refinement~\cite{vázquez2024llms}. 
However, \revision{these evaluations do not present results for multiple LLMs and focus on the visual artifacts produced by these LLMs.} Our work builds on this thread of research by evaluating the strengths and limitations of different LLMs in inferring the semantic information needed to create visualizations. 

%% file: Sections/ground-truths.tex
\section{Collating Natural Language Utterances}
To facilitate the evaluation of LLMs' capabilities for extracting relevant data and visual contexts, we need a set of data-related user utterances to provide as prompts to LLMs. These utterances need to reflect the level of uncertainty found in human speech. To this end,
we sourced utterances from two publicly available corpora:

\begin{itemize}[nosep]
    \item \textbf{NLVCorpus:} This dataset presents 893 utterances collected from an online survey, where 102 respondents were asked to describe utterances they would input to an analytical system to generate a specific visualization~\cite {srinivasan2021collecting}.
    \item \textbf{Quda:} This dataset utilizes interviews with expert data analysts to generate a corpus of 920 utterances~\cite{fu2020quda}. These utterances were refined and paraphrased via a crowdsourced study to generate a final dataset of 14,035 diverse utterances.
\end{itemize}

\noindent We performed a systematic examination of utterances from each dataset and filtered out utterances if they contained SQL pseudo code, e.g., \textit{``group (region) | For each region, group by (ship status) | For each (region, ship status), calculate the sum of profit''.} For our analysis, we were interested in examining how well LLMs infer the necessary aspects of the semantic profile and not explicit visualization descriptions. Consequently, we also filtered out utterances that specified visualization types or mapping of data to visual elements, e.g., \textit{``give me a scatterplot of imdb rating as x axis and rotten tomatoes rating as y axis''}. 

This selection process was first applied to the NLVCorpus dataset, which yielded a total of 134 utterances across 3 unique datasets. We then applied the same inclusion criteria to a subset of the Quda dataset to produce the remaining 309 utterances across 32 datasets. We also included 54 utterances across 2 datasets collected from a classroom activity \revision{conducted in an undergraduate level data visualization class at a US-based University}. Our final corpus consists of 500 diverse utterances across 37 unique datasets.

%% file: Sections/experiment.tex
\section{Generating Ground Truths and LLM Responses}
\label{corpus:gt}

\subsection{Manually Annotating Utterances}
\label{corpus:gt:annotations}
\revision{Three of the authors performed manual annotation of utterances in our corpus. The lead annotator has 5 years of visualization research experience, while the remaining two annotators have at least 2 years of experience creating visualizations.} To annotate our corpus of utterances, the lead author drafted an initial codebook from an evaluation of relevant taxonomies for visual tasks and data transformations~\cite{amar2005lowlevel, munzner2014visualization}. Five random utterances were then selected from the corpus, and three of the authors independently examined and annotated them. The authors met in a subsequent meeting to discuss their codes. The codebook was then updated based on this discussion. The three authors manually annotated the remaining 495 utterances over the course of 12 weeks, holding weekly meetings to discuss and resolve conflicts. 
Here, we describe these annotations.

\vspace{1mm}
\noindent\textbf{Uncertainties.}
We identified utterances that could lead to multiple interpretations or couldn't be answered based on the provided dataset. We annotated ambiguities and under-specification by highlighting confusing words, explaining their lack of clarity, and suggesting resolutions. For instance, the utterance \textit{``in what manner are good air quality records dispersed throughout the monitored region ?''} was labeled ambiguous because the reference dataset had air quality readings generated at different times for each region. Therefore, the good air quality readings could be split into different time periods (per hour of the day, per date) or even aggregated across the entire dataset. We provided a resolution to calculate summary statistics and generate yearly trends for good air quality.

While annotating the 500 utterances in our corpus, we found 18 utterances that 
requested information unavailable in the dataset. For instance, on the dataset showing life expectancy by states in the US, one of the utterances asked \textit{``show me the GDP ranking of European countries''}. This dataset did not contain any information about any countries. As such, it is not possible to answer such a question. Since these utterances were obtained from other studies, it is unclear how these utterances came to be. While we did not provide annotations for the relevant data and visual context for these utterances, we still chose to include them when prompting LLMs as we are still interested in evaluating their ability to identify and resolve such uncertainties in utterances.

\noindent\textbf{Data Attributes and Transformations.}  
For each utterance, we identified the relevant data column[s] needed to correctly answer the utterance. 
Some utterances require data transformations to generate a new data table that can be used to answer the question. We initially captured the operations needed to transform the data table, such as fold, unstack, and group. However, to properly evaluate if these operations are accurate, we need to evaluate the actual data tables that are generated from these operations. As such, we opted to capture the relevant pandas code that would be used to perform data transformations. Using the previous example utterance on the air quality dataset, the data transformation needed to generate the relevant data table was $res=df.groupby(['Generated', 'Station']).apply(lambda\:x: x[x['Air\_Quality'].lower() ==\:'good'])$

\vspace{1mm}
\noindent\textbf{Visualization Tasks.}
The visual task[s] were classified
based on the inferred intent of the utterance. The taxonomy for these tasks was adopted from published works by Amar et al. ~\cite{amar2005lowlevel} and Munzner~\cite{munzner2014visualization} and include: \texttt{\small Retrieve Value, Filter, Compute Derived Value, Find Extremum, Sort, Determine Range, Characterize Distribution, Find Anomalies, Cluster, Correlate, summarize, Compare, Dependency, Similarity, and Trend}. 

\subsection{Generating LLM Outputs}
We evaluated two proprietary and two open-source LLMs. 

\vspace{.05mm}
\noindent\textbf{Proprietary LLMs.} We evaluated OpenAI's GPT4-Turbo \gpt~\cite{openaigpt42023} and Google's Gemini-Pro \gemini~\cite{gemini}. GPT4-Turbo has a training data cutoff of December 2023 and Gemini-Pro's training data cutoff is described as ``early 2023''~\footnote{According to \href{https://ai.google.dev/gemini-api/docs/models/gemini}{Google AI documentation}}.  We utilized the Application Programming Interfaces (APIs) for both of these models to generate responses for the 500 utterances in our corpus. 

\vspace{0.5mm}
\noindent\textbf{Open Source LLMs.}
We evaluated two open-source LLMs, Llama3 \llama, and Mixtral \mixtral, on the Llama factory code base~\cite{zheng2024llamafactory}. Llama3~\cite{llama3modelcard} has 70 billion parameters and a context length of 8,000 tokens, with a knowledge cutoff of December 2023. Mixtral-8x7B-Instruct~\cite{jiang2024mixtral} is configured with 46.7 billion parameters and similarly has a knowledge cutoff in December 2023.


\subsubsection{Prompt Design}
\revision{We explored different prompting strategies (One-shot vs. Few-shot) to elicit responses from LLMs.} We decided to use a few-shot prompting 
as it is more suited for complex tasks and allows the model to learn requirements from provided examples~\cite{wei2022chain}. The prompt provided to each model contained similar instructions to those used by our human annotators in Sec.\ref{corpus:gt:annotations}. 
For the data transformation code, we also instructed the LLMs not to include code for plots or complex analyses. We also included three utterance-dataset-output samples, which were not included in our evaluation corpus. We chose to include the first 10 rows of the dataset to provide an overview of the input data schema. We also included the corresponding ground truth annotations for our sample utterances to help the model gain an understanding of the expected output. Due to space considerations, the full prompt has been provided in supplementary materials~\footnote{\href{https://osf.io/j342a/wiki/Prompting LLM Scripts/?view_only=b4051ffc6253496d9bce818e4a89b9f9}{Supplementary Materials}}.





\subsubsection{Challenges Retrieving Responses.}
We expected to receive a total of 2000 LLM responses (500 per LLM). However, \revision{we encountered some issues eliciting responses from the LLMs. Some of our queries using the APIs of proprietary models returned null responses (\gpt: 9, \gemini: 2).}  For the open-source models, 42 of the responses did not return the JSON annotations and instead returned a \revision{text-based} answer to the utterance (\llama:20, \mixtral:22). Both models also occasionally failed to correctly format the JSON responses correctly, wrapping keys with `/,` `@,` or `\textless.` Wrongly formatted JSON responses were resolved manually. 
\revision{The final set contains 1947 valid annotations from the LLMs (\gpt: 491, \gemini: 498, \llama: 481, \mixtral: 477).}

%% file: Sections/results.tex
\section{Analysis and Results}
\label{results}
We analyzed the LLMs responses across three dimensions of semantic profiling: 
\textit{clarity analysis} (i.e., comprehension of utterances in the presence of uncertainty), 
proper identification of the \textit{relevant data context}, and proper inference of the \textit{visualization task}.

\subsection{Identifying uncertainty}
\label{results:uncertatinty}

\noindent\textbf{Summary Statistics:} Of the 500 utterances in our corpus, the human annotations found uncertainty in 96 of the utterances. A total of 813 uncertainties were found across all LLMs (\gpt: 268, \gemini: 192, \llama: 180, \mixtral: 173). Of these 813 uncertainties, only 25.1\% (n=204) overlapped with human annotations (\gpt: 74, \gemini: 46, \llama: 44, \mixtral: 40). 

\begin{figure}[t!]
  \centering
  \includegraphics[width=\linewidth]{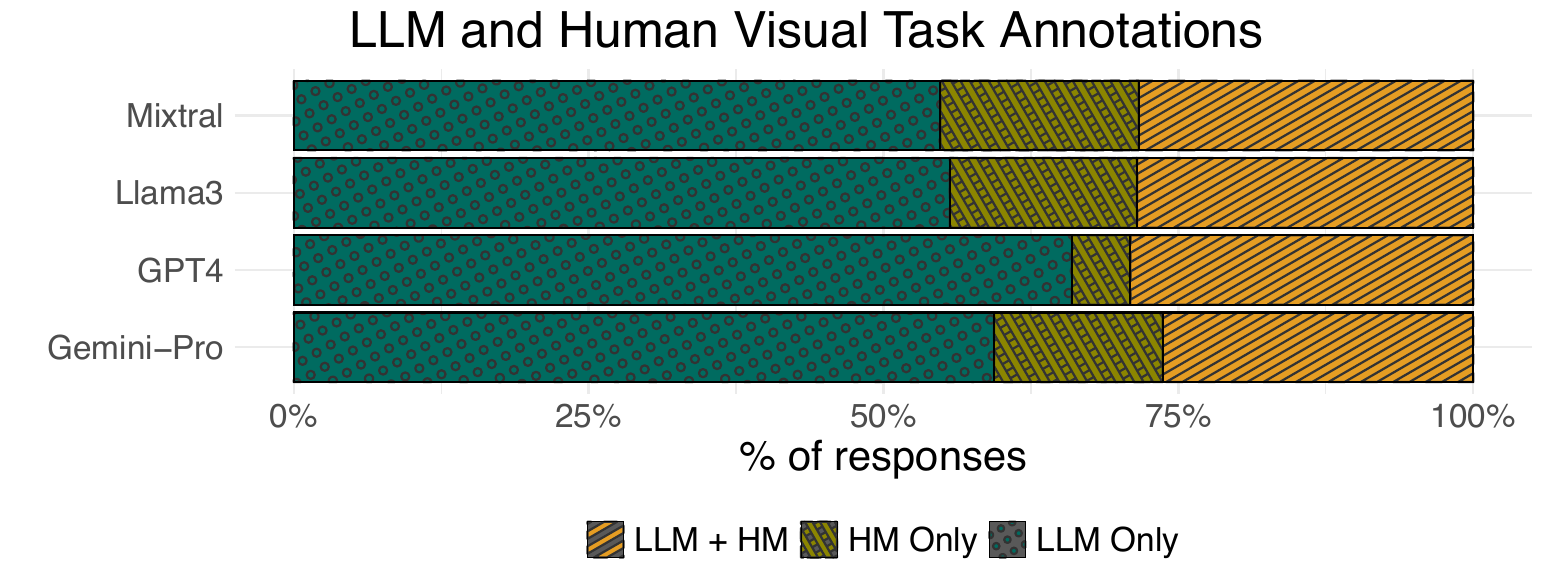}
  \vspace{-5mm}
  \caption{
  Overview of the overlap in uncertainty annotations between the LLMs and Human (HM) annotations.
  }
  \vspace{-3mm}
  \label{fig:abiguity-agreement}
\end{figure}

\vspace{1mm}
\noindent\textbf{Differences in uncertainties classified by LLMs and human annotators.} We observe that all LLMs identified a higher proportion of uncertainty in the utterances than those identified by the human annotators (see Fig.\ref{fig:abiguity-agreement}). When we examine some of these uncertainties identified by the LLMs, we find that they describe uncertainty on how to perform analysis or missing context for data column values. For instance, for the utterance \textit{``Can we conclude that higher happiness comes from higher freedom?''}, GPT-4\gpt ~returned the following ambiguity: \textit{``The query does not specify if the analysis should consider other factors that might influence happiness, or if it should be isolated to just happiness and freedom.''} To the human annotators, this was simply a case of showing the correlation between the two attributes; hence, there was no uncertainty annotation for this utterance. Similarly, for the utterance \textit{``Compare the number of tall buildings in Hong Kong with Taiwan''},  Gemini-Pro\gemini~ classified this as uncertain because \textit{``It is unclear what metric should be used to quantify the tallness of a building. Should the number of stories be used or the height in meters or feet?''}. Our human annotators inferred that the height of the building would be the measure used to answer this utterance. 

\vspace{0.5mm}
\noindent\textbf{Uncertainties not found by LLMs.} Of the 96 utterances for which human annotators found uncertainty, some were not identified by LLMs (\gpt: 14, \gemini: 32, \llama: 34, \mixtral: 35). A majority of these uncertainties were as a result of either missing or conflicting data being referenced in the utterance. An example is the utterance ``\textit{How can the population of Ashley be illustrated to show the distribution across five years}?'' Our annotations labeled this as uncertain because the dataset only contains information from 2000 to 2002, so it is impossible to answer this using the dataset. None of the LLMs labeled this utterance as uncertain. 
\begin{figure*}[t!]
    \centering 
\begin{subfigure}{0.3\textwidth}
  \includegraphics[width=\textwidth]{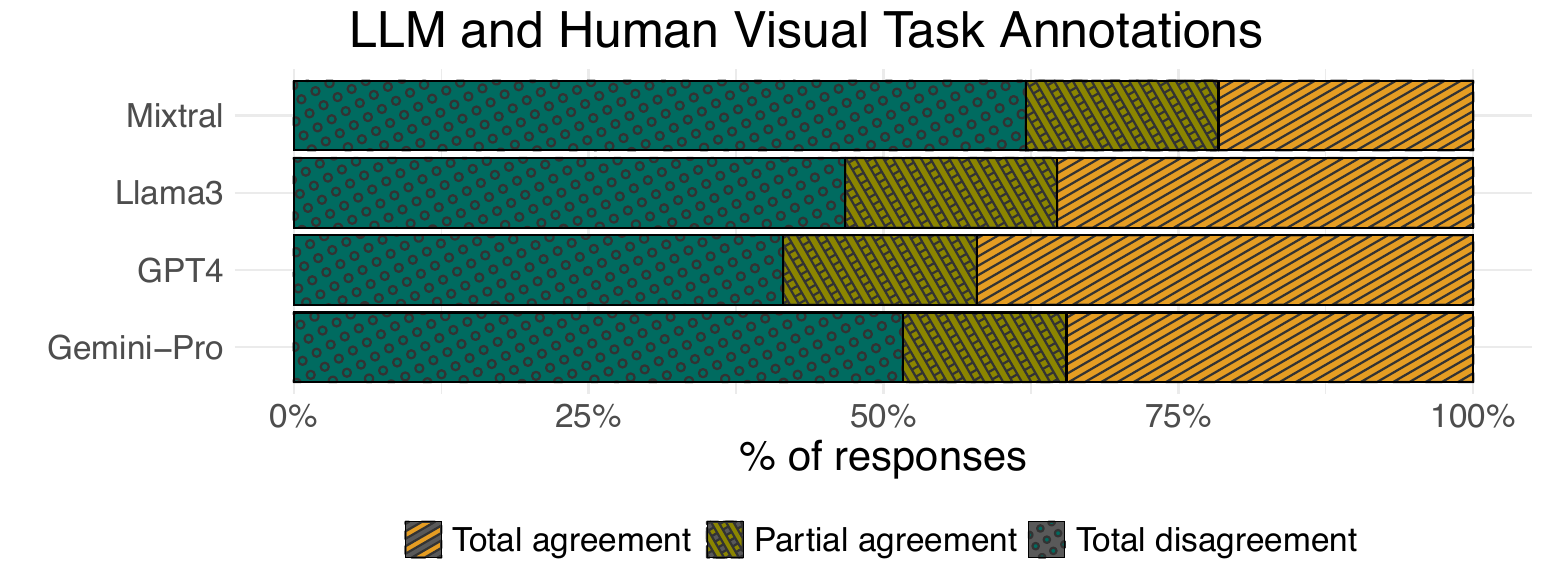}
  \vspace{-4mm}
  \caption{
  Agreement between LLM and human annotations for relevant data columns.
  }
  \vspace{-2mm}\label{fig:agreement-data}
\end{subfigure}\hfil 
\begin{subfigure}{0.3\textwidth}
  \includegraphics[width=\textwidth]{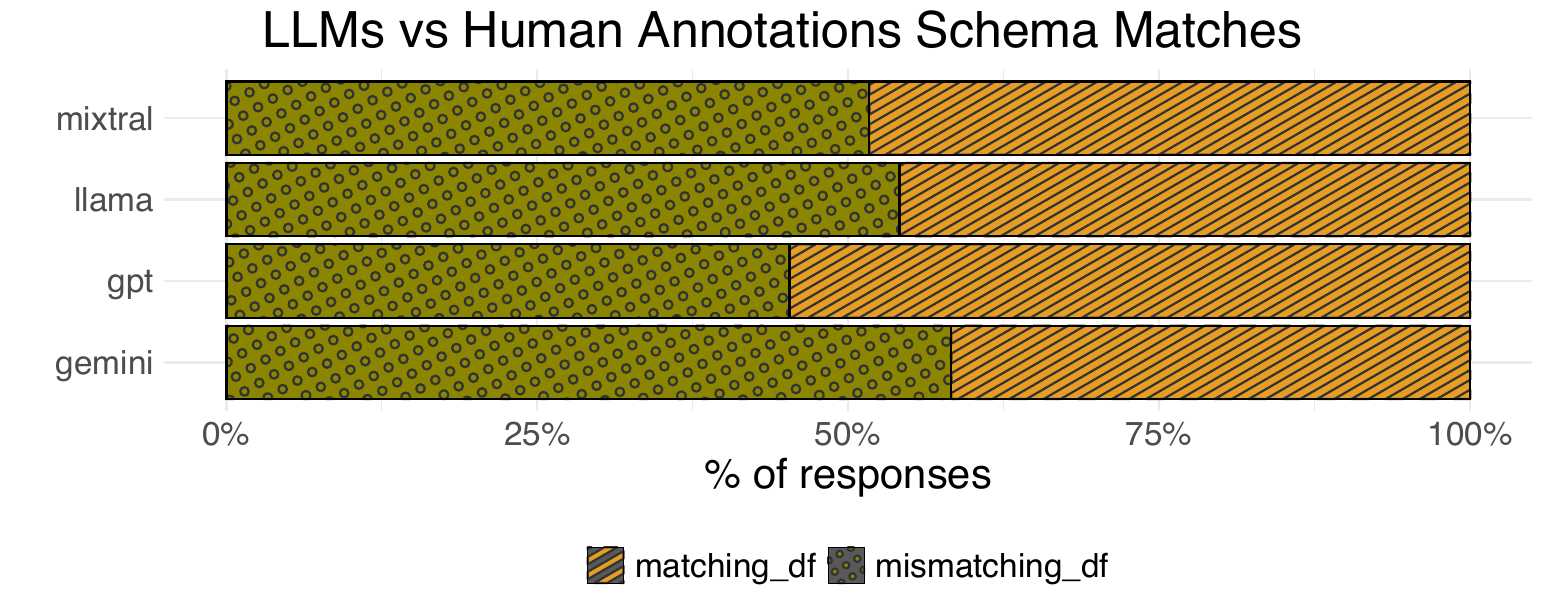}
  \vspace{-4mm}
  \caption{ Data schema matches between data tables returned by LLMs generated code and human annotations.
  }
  \vspace{-2mm}\label{fig:data-transforms}
\end{subfigure}\hfil 
\begin{subfigure}{0.3\textwidth}
  \includegraphics[width=\textwidth]{figures/agreement_vis_task_pattern_fill.pdf}
  \vspace{-4mm}
  \caption{
  Agreement between LLM and human annotations for visualization tasks.
  }
  \vspace{-2mm}\label{fig:agreement-vis-task}
\end{subfigure}
\caption{Overview of overlapping annotations between LLMs and humans for data attributes, transformations and visual tasks.}
\label{fig:data-vis-figs}
\end{figure*}

\subsection{Identifying Relevant Data Context}
\label{results:data-context}
For each data column identified in LLM-generated responses, we examined if they were also identified by human annotators. We defined three levels of agreement between LLMs and human annotations: 1) \textit{total agreement}, where LLMs identify all relevant data columns; 2) \textit{partial agreement}, where LLMs identify some of the data columns; and 3) \textit{total disagreement}, where LLMs identify none of the data columns.

\vspace{0.5mm}
\noindent\textbf{Summary Statistics.} Of the 1947 responses returned by LLMs, we filtered out 53 responses that were related to the utterances for which our human annotators did not generate codes for data columns (see Sec.~\ref{corpus:gt:annotations}). We also eliminated an additional 13 responses where the LLMs did not generate data column values, bringing the total responses evaluated for data columns to 1881. 

\vspace{0.5mm}
\noindent\textbf{LLMs are able to correctly infer relevant data columns for most utterances.} \revision{As shown in~\cref{fig:agreement-data}}, 57.5\% of the valid annotations generated by LLMs had a total agreement with the human annotations (\gpt:312, \gemini:241, \llama:273, \mixtral:255). 34.24\% had partial agreement(\gpt:140, \gemini:180, \llama:157, \mixtral:167) between LLMs and human annotations, while 8.29\% had complete disagreement in the relevant data columns identified (\gpt:32, \gemini:48, \llama:37, \mixtral:39). We observed that 43.6\% of these complete disagreement cases 
had uncertainties identified by either human annotators or LLMs.

\subsubsection{Data transformations}
\label{results:data-context:transforms}
For each response generated by an LLM, 
we executed both the LLM-produced and human-annotated transformations, extracted the resulting data tables from both executions and compared their underlying data schemas (i.e., attribute types) to verify the accuracy of the transformations presented by LLMs. For example, for the utterance \textit{``What is the relationship, if any, between wind and pressure?''}, both the data transforms provided by Llama3\llama~ and human annotations returned a data table with the following schema \{${wind: int64, pressure: int64}$\}. Since the data tables have the same number and types of attributes, this is a positive match.

While evaluating the data transformations, we found  31 instances where the code for data transformations violated instructions on not returning code for visualization plots or performing complex analyses which were excluded from our analyses (\gpt:1, \gemini:0, \llama:15, \mixtral:15). Furthermore, we found that 385 of the transformations raised errors of various kinds (\gpt:59, \gemini:96, \llama:119, \mixtral:111) or returned raw values and not data tables (\gpt:66, \gemini:90, \llama:57, \mixtral:52). Since the human annotation prioritized data tables as the output of data transformations, we exclude such responses in our analyses.


\vspace{1mm}
\noindent\textbf{Data transformations produced by LLMs do not always match those generated by human annotators.} The final set for our analysis on data transformation is 1238 responses (\gpt:360, \gemini:290, \llama:292, \mixtral:296). 48.1\% of these responses produced data tables with schemas that match those produced by the human annotations (see \cref{fig:data-transforms}). For the remaining 51.9\% where the data did not match what was produced by the code annotated by humans, our evaluation focuses on matches between data schemas. As such, we cannot verify if the resulting data tables provide meaningful answers to the utterance or if they were the result of incorrect data transformations.


\subsection{Inferring Visualization Tasks}
\label{results:vis-context}
Similar to the analysis for data columns, we identify three levels of agreement between human and LLM annotations for visual tasks.

\vspace{0.5mm}
\noindent\textbf{Summary Statistics.} Of the 1947 responses returned by LLMs, visualization tasks were identified in 1940 responses (\gpt:490, \gemini:494, \llama:479, \mixtral:477).


\vspace{0.5mm}
\noindent\textbf{Higher proportion of disagreements between human annotations and LLMs for visual task classifications.} We observed \revision{the highest} level of disagreement between LLMs and human annotations in the visual task classifications. \revision{50.4\% of the visual tasks were in total disagreement, as seen in Fig.~\ref{fig:agreement-vis-task} (\gpt:205, \gemini:253, \llama:224, \mixtral:296). There was total agreement in 33.43\% of the responses (\gpt:208, \gemini:169, \llama:169, \mixtral:103) while the remaining 16.17\% had partial agreement for the visual task (\gpt:81, \gemini:68, \llama:86, \mixtral:78).} When we examine a portion of the cases with total disagreement, 
we observe that some of the issues are a result of conflicting interpretations. For instance, for the utterance \textit{``What is the main factor depending on different status (wind, time, pressure, etc)?''}  Gemini\gemini~ classified this as ``correlation'' whereas the human annotations classified the utterance as ``dependency'' since correlation cannot be calculated between categorical and numerical attributes.  We also see instances where LLMs mix data transformations with visual tasks, e.g., for the utterance \textit{``What was the average budget for each content rating and creative type, as multiple column charts?''} Mixtral\mixtral~ classified the utterance as ``aggregation, categorization \& relationship''.




%% file: Sections/discussion.tex
\section{Discussion And Future Work}
We evaluated the capabilities of four publicly available LLMs in semantic profiling of natural language utterances for data visualization. Our results pose interesting insights for future research.

\vspace{0.5mm}
\noindent\textbf{Using uncertainties to facilitate deeper data exploration and analysis.} Our findings show that LLMs found a higher number of uncertainties in utterances compared to our human annotators. It is possible that humans and LLMs identify uncertainties at different levels of abstraction, as humans are able to interpret context more deeply and make better inferences. One such instance of this difference can be seen in the inference in the ``tallest building'' example provided in 
 Sec.\ref{results:uncertatinty}. As a result, LLMs might be more sensitive to uncertainties in utterances. However, this may not be a limitation as their sensitivity to uncertainties can be leveraged to pose questions to analysts and help them think deeply about their analysis questions or approach. Facilitating such interactions in NLIs is an interesting research direction.

\vspace{0.5mm}
\noindent\textbf{Improving programming-based responses to utterances.} We observed that LLMs are also capable of inferring the appropriate data columns and transformations for over half of the utterances. Yet, for many of the data transformations, we found a number of issues within the code returned by LLMs. This issue is known and tools circumvent this by prompting for multiple code scripts and filtering out erroneous scripts~\cite{dibia2023lida, wang2024dataformulator}. While these erroneous responses can improve via feedback and fine-tuning prompts, there is a need for further research on how to improve the generation of relevant code for visualization contexts.

\vspace{0.5mm}
\noindent\textbf{Improving visualization task inference to facilitate exploration.} We also found that LLMs struggle to correctly infer appropriate visualization tasks from utterances. 
Nevertheless, there is a need to investigate ways to improve LLMs' ability to infer visualization tasks properly. This is important as these tasks often inform visualization design choices, such as using bar charts for comparison or violin plots to characterize distributions~\cite{amar2005lowlevel, munzner2014visualization, narechania2021nl4dv}. Proper inference of visual contexts can also facilitate a breadth-wise exploration of data similar to the Voyager system~\cite{wongsuphasawat2016voyager}. For instance, if a user is working on the movies dataset and an LLM can infer they are trying to \textit{find anomalies} in the IMDB ratings, it can recommend potentially interesting utterances based on the relevant tasks, 
such as \textit{comparing} IMDB ratings across creative tasks or finding \textit{correlations} between IMBD and Rotten Tomato ratings. 

\section{Conclusion}
We evaluated the capabilities of four publicly available
LLMs (GPT-4\gpt, Gemini\gemini, Llama3\llama~and Mixtral\mixtral) at correctly inferring the semantic profiles of natural language utterances for data visualization generation. Our findings reveal important strengths of LLMs at identifying uncertainties in utterances and inferring relevant data columns. We also highlight the current limitations of LLMs for generating data transformation code and inferring visualization tasks. Based on our findings, we present future research directions on the use of LLMs for visualization generation. 